\documentclass{article}
\usepackage{ijcai18}

\usepackage{times}
\usepackage{xcolor}
\usepackage{soul}
\usepackage[utf8]{inputenc}
\usepackage[small]{caption}

\usepackage{graphicx}
\usepackage{dblfloatfix}

\usepackage{tabularx}
\usepackage{tabulary}

\usepackage{url}
\title{Deep Learning Based Multi-modal Addressee Recognition in Visual Scenes with Utterances}

\author{
Thao Minh Le$^1$ \thanks{This study was conducted during an internship at Yahoo! JAPAN Research, Tokyo, Japan}, 
Nobuyuki Shimizu$^2$, 
Takashi Miyazaki$^2$, 
Koichi Shinoda$^1$ 
\\ 
$^1$ Tokyo Institute of Technology, Tokyo, Japan \\
$^2$ Yahoo Japan Corporation\\
thao@ks.cs.titech.ac.jp, 
\{nobushim, takmiyaz\}@yahoo-corp.jp,
shinoda@c.titech.ac.jp
}

\begin{document}

\maketitle

\begin{abstract}
With the widespread use of intelligent systems, such as smart speakers, addressee recognition has become a concern in human-computer interaction, as more and more people expect such systems to understand complicated social scenes, including those outdoors, in cafeterias, and hospitals. Because previous studies typically focused only on pre-specified tasks with limited conversational situations such as controlling smart homes, we created a mock dataset called Addressee Recognition in Visual Scenes with Utterances (ARVSU) that contains a vast body of image variations in visual scenes with an annotated utterance and a corresponding addressee for each scenario. We also propose a multi-modal deep-learning-based model that takes different human cues, specifically eye gazes and transcripts of an utterance corpus, into account to predict the conversational addressee from a specific speaker's view in various real-life conversational scenarios. 
To the best of our knowledge, we are the first to introduce an end-to-end deep learning model that combines vision and transcripts of utterance for addressee recognition.
As a result, our study suggests that future addressee recognition can reach the ability to understand human intention in many social situations previously unexplored, and our modality dataset is a first step in promoting research in this field.  
\end{abstract}

\section{Introduction}
	\label{sec:introduction}
Building human-friendly robots that are able to interact and cooperate with humans has been an active research field in recent years. A major challenge in this field is to develop intelligent systems that can interact and cooperate with people outside a laboratory setting. Applications include a robot companion acting as an assistant, such as a robot butler, mechanical seeing-eye dog, robot lifeguard, and mobile nursing care robot. One can easily imagine how useful a robot butler can be if it can oversee family occasions, such as home parties, and cater to various needs of guests and family members.

We believe that the next generation of intelligent systems will need to possess the ability to understand people in a scene outside a laboratory, from their utterances, gazes, postures, and how they interact with the scene and the others around them. To be truly communicative, an intelligent system needs to detect if a person is initiating an interaction with it and hold a meaningful conversation with people in natural language. However, prior studies on addressee recognition have faced the following problems.

First, studies on addressee recognition and detection have focused on simple pre-specified tasks such as playing games, guiding art experiences, and controlling smart homes, and have not taken into account the diversity of conversational situations.
For example, the vernissage corpus \cite{Jayagopi:2013:VCC:2447556.2447611} has 13 sessions of a robot interacting with two people in an office setting, lasting around 11 minutes. These studies used a Wizard-of-Oz method to manage the dialog as well as the robot's gaze and nodding. 
Because their scenarios involved a stationary robot, they were able to obtain video and audio recordings and other sensory data. The downside is the limited variations in situations in which the conversations occur. Other studies, such as \cite{HOLTHAUS16.1046}, assumed a smart home setting. Again, their conversational situations were very specific and fixed. 
  
Second,  previous studies have mainly focused on gaze and non-verbal information \cite{vanTurnhout:2005:IIA:1088463.1088495}, including head pose \cite{johansson2013head}.  
However, in many social situations, gaze and head pose may not be sufficient for addressee recognition and detection. For example, in the case in which joint attention or shared attention (the shared focus of two individuals on an object) occurs, the speaker may not be looking at the addressee when he/she speaks. Thus, the textual content of utterances may also play a major role in addressee recognition \cite{akhtiamov2017speech,tsai2015study,Tsai2015MultimodalAD}. 
Combining both visual scene information and spoken utterances for addressee recognition is difficult, which is what Sheikhi and Odobez \shortcite{sheikhi2015combining} attempted. Their experiments were limited to laboratory dialog context in which a computer agent acted as an art exhibition guide interacting with two humans and involved sophisticated graphical models. In other words, the conversational context in the experiments was only focused on interactive question and answering regarding a few pre-selected artworks as well as general art topics. By increasing the variety of conversational contexts, we expect to mitigate the limitations of previous studies in this regard. Although both visual features and utterances are helpful in addressee recognition, studies that use both features have just started to appear in the literature. We believe much more work needs to be done in multi-modal addressee recognition.

To address the two concerns above, we created a mock dataset using a pre-existing image dataset with large variations in visual scenes by further annotating utterances and addressees using crowdsourcing. To attend to a large number of social situations a robot may face, we determined a trade-off. We have forgone audio and video components of previous studies and opted to increase the variations in social situations by creating mock conversations instead of conducting Wizard-of-Oz experiments. The end result was the creation of our Addressee Recognition in Visual Scenes with Utterances (ARVSU) dataset. As the image dataset we used contains gaze annotations, it contains sufficient information to kick start addressee recognition research in a large number of social situations. Compared to previous Wizard-of-Oz studies, our corpus is focused on more day-to-day encounters. Examples include people talking about food in a cafeteria, posing for a photograph in front of a historic site, and watching a baseball game and talking about it.

We also propose a multi-modal, deep-learning-based model for addressee recognition incorporating gaze and utterance. Although there have been studies that used deep learning models for addressee detection using speech, to the best of our knowledge, we are the first to introduce an end-to-end deep learning model that combines vision and transcripts of utterances for addressee recognition. We also empirically show that exploiting both utterances and images improve the overall performance of the addressee recognition system.

The paper is organized as follows.
In Section \ref{sec:previous_studies}, we discuss the related studies.
In Section \ref{sec:dataset}, we introduce our dataset, including its construction and statistics. 
We then explain our proposed model in Section \ref{sec:method}. 
We explain the experimental results in Section \ref{sec:exp_disc}.
Finally, we conclude the paper in Section~\ref{sec:conclusion}.

\section{Related Studies}
	\label{sec:previous_studies}
\subsubsection{Gaze Prediction}
Human gaze is determined as a crucial human cue used to provide visual interaction and a means of communication in face-to-face conversations \cite{jovanovic2006addressee}. Thus, gaze information is considered beneficial for addressee recognition. We can think of two settings for the problem of gaze prediction based on an image. One is predicting the eye-fixation points from which an observer (or photographer) is looking at an image. The other is predicting the location at which a person in an image is looking. We call the latter the gaze-following problem.

\cite{recasens2015they} constructed a dataset (the GazeFollow dataset) for the gaze-following problem and proposed a deep-learning-based gaze-prediction model. The images of the dataset were gathered by concatenating several major image datasets (SUN, MS COCO, Actions 40, PASCAL, ImageNet detection challenge, and Places dataset). This concatenation results in a large image collection of 130,339 people in 122,143 images performing diverse activities in many everyday scenarios. They also created annotations of the eye and gaze locations of a person in each image by crowdsourcing. They proposed a model to predict the gaze-fixation point of the specified person in the image from the full image, head image of the person, and relative head location in the head image. 

The focus of their work was finding the direction of gazes, which is quite different from our focus on addressee recognition. 
However, in face-to-face conversation, the gaze tends to focus on the addressee. Thus, we decided to exploit pre-existing gaze annotations in shaping our dataset. 
Because we also postulate that utterances may play an important role in addressee recognition, we have taken the GazeFollow dataset and annotated likely utterances of people in images and to whom they may be addressed using the crowdsourcing service Amazon Mechanical Turk. 

\subsubsection{Deep Learning for Addressee Recognition}
Although deep learning outperforms traditional methods to achieve the state-of-the-art results in most topics in artificial intelligence, studies on addressee recognition by applying deep learning is limited. Published studies following this direction mainly used deep learning models, specifically recurrent neural networks (RNNs), for time-sequence data. \cite{pugachev2017deep} applied different deep learning techniques, including a fully connected deep neural network (DNN) and a bidirectional long short-term memory (BLSTM), for an acoustic modality. While the DNN surpassed a traditional classifier in their study, BLSTM performed poorly for this task. In contrast, \cite{ouchi2016addressee} generated a simulated multi-party conversation corpus then used an RNN as an encoder model for lexical content extraction.

Our study, on the contrary, was focused on applying extra-linguistic information, which is recognized as equivalent to the speech signal in a spoken-dialog system with 100\% confidence, for an addressee-recognition system using multi-modal data. With this intention, we leverage the power of crowdsourcing services to create a linguistic corpus on daily-life conversations from the speakers' side. At the same time, we also use the saliency of visual information extracted using a convolutional neural network (CNN) to estimate human-eye gaze and use the estimated human gaze together with the speakers surrounding environments to analyze a mock conversation.

\subsubsection{CNN and Transfer Learning}
The image-recognition performance of CNNs has advanced rapidly in recent years. As a result, CNNs are now widely used for various image-recognition tasks as well as other related tasks. However, training a CNN from scratch on a very large dataset is extremely time consuming and computationally expensive.
Recently, it has been more common to take a fully trained model on a large-scale dataset and fine-tune it on a new dataset for a specific task, which is known as transfer learning \cite{pan2010survey}. 
Transfer learning is known to be very powerful and has been widely applied in various applications such as object recognition, segmentation, and retrieval
\cite{oquab2014learning}
as well as scene classification \cite{donahue2014decaf}. 
 Similar to these studies, we also used ImageNet and VGG-Face pre-trained CNNs as feature extractors for visual feature representation.

\section{Dataset}
	\label{sec:dataset}
As noted in Section 1, to model interactions between people and a robot not limited to indoor settings, we have opted for creating a mock dataset. In this section, we describe the data statistics and how we gathered the data for our ARVSU dataset. 

\subsection{Scenario Description}
	\label{section:scenario}
\begin{figure}[!t]
\centering
\includegraphics[width=\columnwidth, clip]{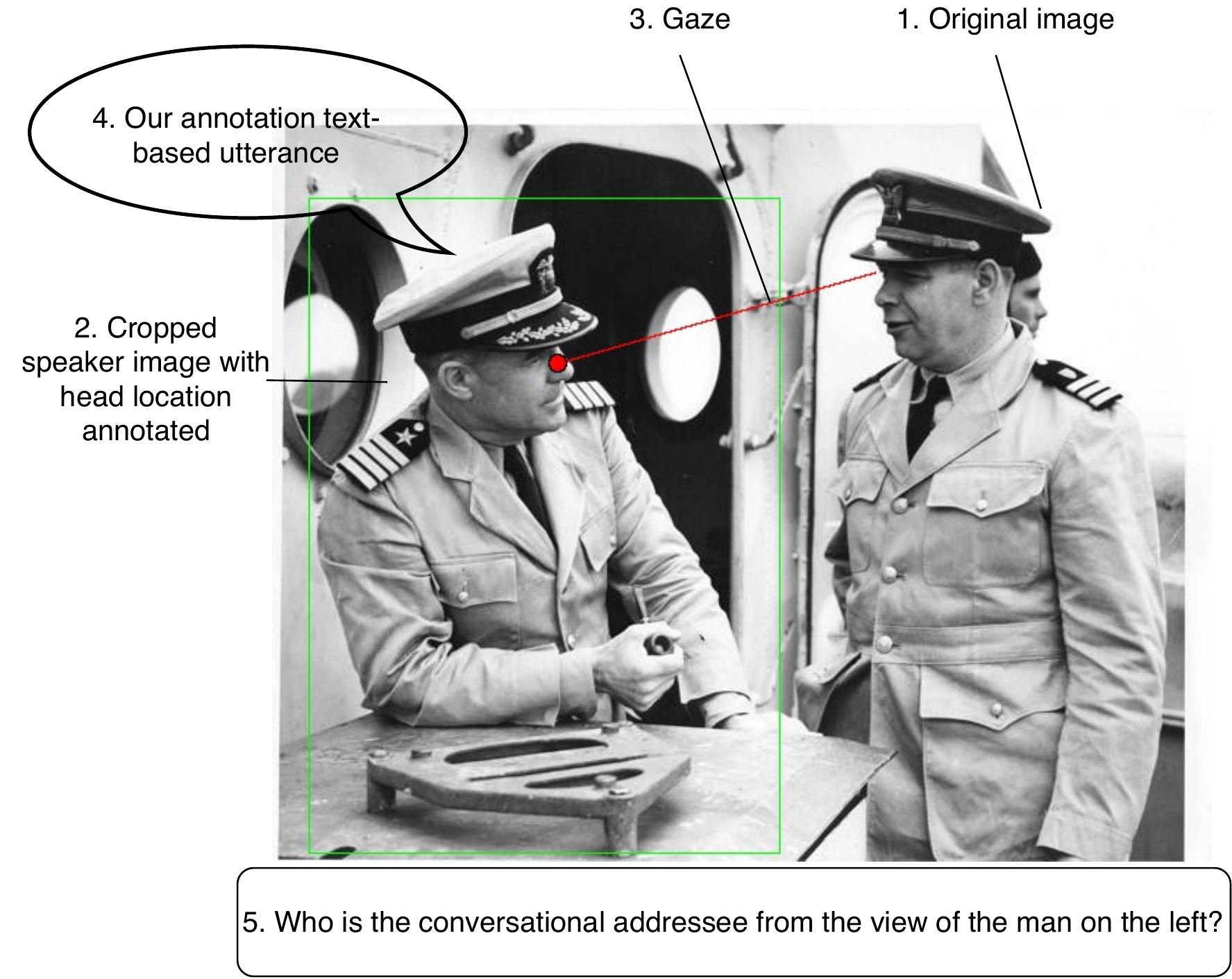}
\caption{Example of Images We Used in Our Data Collection}
\label{fig:sceillus}
\end{figure}
As shown in Figure \ref{fig:sceillus}, our dataset contains images and annotations from the GazeFollow dataset \cite{recasens2015they}. The GazeFollow dataset consists of (1) the original image, (2) cropped speaker image with head location annotated, and (3) gaze. To create our dataset, we further annotated (4) utterances in texts, and (5) to whom an utterance is addressed. The objective of our research was to predict (5) the addressee from the rest of the inputs (1)-(4). In this scenario, a computer agent plays the role of a photographer taking a snapshot of a conversation (the original image). In other words, given an image in our dataset, the task for an intelligent system is to assume a photographer's point of view and recognize who the addressee is as the photographer corresponds to a robot in a multi-party conversation scenario.

Our ARVSU dataset will be released at \url{https://research-lab.yahoo.co.jp/en/software/}.

\subsection{Generation of Simulated Utterances}
	\label{sec:mec_turk}

To create our dataset, we used Amazon Mechanical Turk, which is a microtask-based crowdsourcing platform in which requesters post microtasks that have questions to be asked and workers perform the microtasks.

To create our dataset, in short, we asked the following question to the crowd workers: ``There is a person marked with a red dot in the image. Please imagine that this person is saying something. If this person speaks something in the context of this image, (1) to whom is he/she speaking and (2) what is he/she likely saying?''

The options for choosing the addressee included the following: 
\begin{enumerate}
\setlength{\parskip}{0cm}
\setlength{\itemsep}{0cm} 
\item[a)] The photographer of the image
\item[b)] People, animals, etc... who are in his/her line of sight
\item[c)] Others
\item[d)] Himself/herself (monologue, pondering)
\item[e)] Not applicable 
\end{enumerate}
The crowd workers were instructed to check all that apply. For the utterances, they were asked to provide a sentence. The difference between options b and c is whether an addressee is in the line of sight. Option c includes other people in or out of the image that are not in the speaker's line of sight. Option e is used to remove images without significant human interaction. The detailed instructions will be released as a part of the dataset documentation.

\subsection{Dataset Analysis}
	\label{sec:data_ana}
As a result of the corpus-generated procedure described in Section \ref{sec:mec_turk}, generated utterances are separated into five distinct classes. Table \ref{tab:table1} shows the statistics of our corpus and dataset in detail. 
\begin{table}[t!]
 \small
  \centering 
    \begin{tabular}{l|r|c}
      \textbf{} & \textbf{No. of Utterances}& \textbf{Percentage (\%)} \\
      \hline
      Line-of-Sight Entities & 322,911 & 46.32 \\
      Photographer & 87,373 & 12.53\\
	  Monologue/Pondering & 165,177 & 23.69 \\ 
      Others & 109,124 & 15.65 \\
	  Not Applicable & 12,528 & 1.80  \\	
      \hline
      \textbf{Total} & \textbf{697,113} & {}\\ 
    \end{tabular}
    \caption{Dataset Statistics}
    \label{tab:table1}
  \vskip\baselineskip
\end{table}
\smallskip
Nearly 50\% of the annotated utterances belong to the case in which the speaker is interacting with entities in his line-of-sight within given visual scenes. In contrast, only 12.53\% of utterances are considered addressing the photographer. The ``others'' case, in which the speaker is interacting with an addressee not in his line-of-sight, accounts for over 15\% of the utterances. More than 25\% of the dataset falls into non-human-computer interaction, including monologue, pondering, and other non-defined cases. 

In our corpus, annotators created addressee labels and utterances at the same time, as we thought that this should improve the dataset’s reliability. Because it is difficult to use utterances for inter-annotator agreement and addressee labels are associated with the utterances, we instead took 1\% of the dataset and double checked the annotation using crowdsourcing. It is important to note that 5.5\% of checked samples had utterances that do not correspond well with the situation in the images. To determine the difficulty of the task, we also asked crowd workers if the speaker in the image is looking at the photographer. The results indicate that when the addressee is the photographer, 89.6\% of the cases were when the speaker is not looking at the photographer. This disagreement between gaze and addressee makes predicting the photographer by only using gaze very difficult. On the other hand, for the non-photographer, only 6.2\% of samples were difficult cases. There is usually only one person in the line-of-sight, but it can be one person or a group of people that are outside the line-of-sight.

To ensure the quality of annotation, we also sampled 100 annotations and manually checked their quality. Out of 100 samples, we found 6 samples with obvious errors in addressee labels. We also labeled addressee labels independently for 100 samples and computed Cohen's Kappa value, which is an agreement coefficient that calculates the level at which annotators agreed on label assignments beyond what is expected by chance \cite{doi:10.1177/001316446002000104}. Our lab members labeled line-of-sight entities and obtained Cohen's kappa = 0.46, a moderate agreement level for observational studies \cite{Hallgren2012,R17-1015}. Other labels had a fair agreement level: photographer with kappa = 0.32, monologue with kappa = 0.34, and others with kappa = 0.29. As about half of utterances are labeled line-of-sight entities, we would say the overall agreement of our corpus is moderate. 
\begin{figure}[t!]
\centering
\includegraphics[width=\columnwidth, clip]{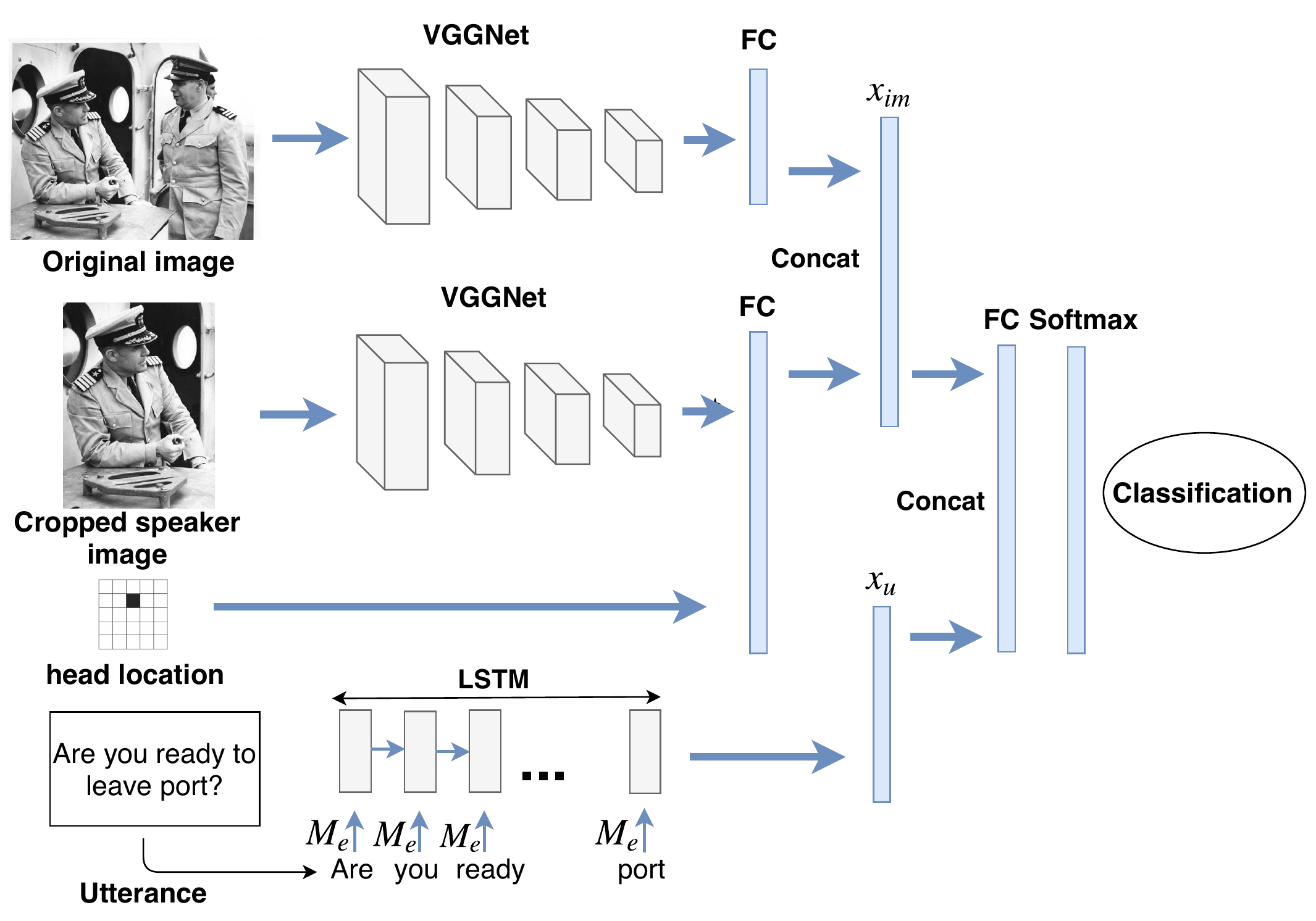}
\caption{Network Architecture}
\label{fig:network_arch}
\end{figure}

Because our dataset was artificially created, it is inevitable to have some potential biases. First, compared to the real world, a disproportionately large number of utterances seem to be labeled as monologue. Labels often co-occur as data are collected in a multi-labeling setting. For example, some annotators could attach two or more labels to an utterance, while others could attach just one. Labels could be difficult to differentiate, especially when line-of-sight seems to go outside the image. For example, a protagonist is seated at a dinner table and having dinner. The image shows only the protagonist and half of the table, and he is facing the other end of the table. His line-of-sight goes outside the image. If we assume that someone is seated in front of him, he would be looking at him/her, so the label would be people. If that is not the case, he may be having a monologue. It is sometimes up to the annotator's imagination to decide what the label would be. In such cases, utterances may be a decisive factor in determining the addressee label, and our corpus to some extent could suffer from a similar problem as with a visual question-answering corpus, where some questions are answerable without looking at the image \cite{VQA}. This problem with line-of-sight going outside the image is a difficulty with the gazefollow dataset annotation as well because the gazefollow dataset always annotates gazes to be within the image, even though people sometimes are obviously looking outside the image.
Despite these limitations, we believe our dataset provides future opportunities to reduce biases, for example, by treating samples with a gaze near the image differently from other images or by having additional line-of-sight classes and more fine-grained addressee classes. 

\section{Multi-modal Addressee Recognition}
	\label{sec:method}
\subsection{Model Overview}
	\label{sec:md_overview}
In this section, we give an overview of our model for addressee recognition in visual scenes, where the inputs of the model are human-gaze information estimated from visual data and annotated utterances. Our proposed model is motivated from intelligent systems that connect natural language processing and computer vision, such as image captioning \cite{vinyals2015show} and visual question and answer \cite{antol2015vqa}. 
Therefore, the network architecture of our proposed model is composed of three main components, as illustrated in Figure \ref{fig:network_arch}: a saliency-estimation-feature stream extracted from original images, speaker-appearance-feature stream extracted from cropped speaker images and annotated head locations of the assumed speakers, and utterance-based-feature stream for understanding the conversational context from lexical information. The two visual-feature streams are used to estimate the human-gaze direction similar to the method proposed by \cite{recasens2015they}. Different from the original gaze-estimation study, we did not use the gaze line annotations available in the GazeFollow dataset. The reason is that while most visual scenes in the GazeFollow dataset only involve interactions within images, we additionally considered interactions with people outside the visible content of the images (photographer).

Mathematically, our learning task is formulated as 
\begin{eqnarray}
  \theta^* = \arg \max_{\theta} p(a|I_1, I_2, S_1, ..., S_{T}; \theta),
\end{eqnarray}
where $a$ is the addressee class, $I_1$ denotes the saliency feature extracted from original images using a pre-trained CNN, $I_2$ depicts the speaker-appearance features obtained by combining the CNN-extracted feature of cropped speaker images and the speaker's head location, and $\theta$ represents the model parameters. The terms $S_1, ..., S_{T}$ denote one-hot vectors of words in an utterance, where $T$ is the number of words in the utterance.

To model $p(a|I_1, I_2, S_1, ..., S_{T}; \theta)$, we one-by-one formulate the model for each modality stream. First, since saliency-estimation-feature and speaker-appearance-feature are followed by a fully connected layer before applying a fusing step by concatenation in the visual-feature streams, the fused output of the two visual-feature streams is formulated as
\begin{eqnarray}
	x_{1}=\textrm{ReLU}( W_{1} I_1+b_{1}) \\
    x_{2}=\textrm{ReLU}( W_{2} I_2+b_{2}) \\
    x_{im} = \textrm{concat}(x_{1}, x_{2}),
\end{eqnarray}
where $W_{1}, W_{2}$ are network parameters, $b_{1}, b_{2}$ are bias values, $\textrm{ReLU}$ is a rectified linear unit function given by $\textrm{ReLU}(x)=\max(0,x)$, $x$ is the input, and "concat" represents a concatenate function. 

Second, regarding text representation, we then use an RNN, specifically a long short-term memory (LSTM) \cite{hochreiter1997long}, which is ubiquitously used in sequence learning. We explain the fundamentals of a CNN for visual-feature extraction as well as an LSTM in the next section. Given one-hot vector representation of word $S_1, ...,S_{t},..., S_{T}$ of an utterance, we first embedded the words into a geometric space $u_t$ by multiplying $S_t$ by an embedding matrix $M_e$ initialized using Global Vectors for Word Representation (GloVe) \cite{Pennington14glove:global}. The embedded vectors $u_t$ are then fed into an LSTM network:
\begin{eqnarray}
	x_{t} = \textrm{LSTM}(u_t, h_{t-1}),
\end{eqnarray}
where $h_{t}$ presents the hidden state at time $t$, and $x_{t}$ is the output of sequence modeling that encodes the semantic of the $t$-th word in the given utterance. Assume that $x_{u}=x_{T}$ is the representation vector of an utterance after the LSTM.

A combination of visual-feature streams and a text-representation stream is then used to model $p(a|I_1, I_2, S_1, ..., S_{T})$
\begin{eqnarray}
	x_{fu} &=& \textrm{concat}(x_{im}, x_{u}) \\
    p(a|I_1, I_2, S_1, ..., S_{T}) &=& \textrm{softmax}(W_{fu}x_{fu}+b_{fu}),
\end{eqnarray}
where $W_{fu}$ and $b_{fu}$ are network parameters and bias values, respectively.
 The term ``softmax'' indicates the softmax function given by
    $\textrm{softmax}_i(v) = e^{v_i}/ \sum_{j} e^{v_j}$,
 where $v$ is a vector.

\subsection{CNN for Visual Feature Extraction}
Our model uses the 16-layer VGGNet \cite{simonyan:deepconv}, which was one of the top performers at the ImageNet Large Scale Visual Recognition Challenge in 2014, as a feature extractor. The 16-layer VGGNet is composed of 13 convolutional layers and 3 fully connected layers. The output of the fc6 of a 4096-dimensional vector is then chosen as the extracted feature vector. In this study, the last two fully connected layers from the original 16-layer VGGNet were discarded. The same process is applied to both original image data and cropped speaker image data. However, to achieve the best performance, we used VGGNet models that were trained on different datasets for the saliency-estimation and speaker-appearance features, which independently play an important role in our model. In particular, we used the VGGNet models trained on the ImageNet and VGG-Face datasets \cite{Parkhi15} for saliency-estimation-feature and speaker-appearance-feature streams, respectively.  

\subsection{RNN/LSTM for Utterance Understanding}
An LSTM is a variant of an RNN that is able to tackle the problem of vanishing and exploding gradients, resulting in the ability to handle longer dependencies compared to a vanilla RNN. 
A typical LSTM is composed of various gates with different responsibilities to control input, output, and the memory behaviors of the network. 
In this study, we used an LSTM with input gate $i_t$, input modulation gate $g_t$, output gate $o_t$, and forgetting gate $f_t$. The number of hidden units $h_t$ was 128. At each time step $t$, the LSTM state $c_t$, $h_t$ was as follows \begin{eqnarray}
    i_t & = & \sigma(W_{ix} x_t + W_{ih}h_{t-1} + b_i) \\
    f_t & = & \sigma(W_{fx} x_t + W_{fh}h_{t-1} + b_f) \\
    o_t & = & \sigma(W_{ox} x_t + W_{oh}h_{t-1} + b_o) \\
    g_t & = & \phi(W_{cx} x_t + W_{ch} h_{t-1} + b_c) \\
    c_t & = & f_t \odot c_{t-1} + i_t \odot g_t \\
    h_t & = & o_t \odot \phi(c_t),
\end{eqnarray}
where $\sigma(x) = (1+e^{-x})^{-1}$ is a sigmoid function, $\phi(x)=(e^x-e^{-x})/(e^x+e^{-x})$ is a hyperbolic tangent function, $\odot$ denotes the element-wise product of two vectors, and $W$ and $b$ are learnable parameters of the network.

\section{Experiments and Discussion }
	\label{sec:exp_disc}
\subsection{Experimental Conditions}
We conducted experiments to demonstrate the effectiveness of our proposed model as well as the benefit of our dataset.
We examined the following three models: two unimodal models {\bf visual information only} and {\bf text-based utterance only}, and our proposed multi-modal model {\bf visual information and utterance}.

The first model uses visual information only; thus, we addressed the problem of addressee recognition by using the gaze information estimated from only visual features. This baseline is used to show the contribution of visual information to understand the conversational context. The second model uses text-based utterance information only; thus, only the utterances were used to understand the conversational context based on the analysis on the semantic meaning of a single spoken transcript. The third model (proposed model) uses visual information and text-based utterance. As described above, our model is composed of visual streams and a text-based-utterance stream that are expected to leverage both visual information and utterances for understanding the conversational context. The implementation strictly follows the network architecture described in Section \ref{sec:md_overview}.

As mentioned in Section \ref{sec:data_ana}, there are five possible addressee classes defined in our dataset, namely, ``Line-of-Sight Entities'', ``Monologue'', ``Photographer'', ``Others'', and ``Not applicable''. However, since our experiments were conducted to address the problem of addressee recognition for simulating a multi-party conversational scenario, we used the ``Photographer'' class as the representation for human-computer interaction and the ``Line-of-Sight Entities'' class for the interaction with a specific entity. As one utterance can be matched with multiple addresses, it is necessary to reorganize the class labels for classification. With this intention, we first removed all cases of the ``Not applicable'' class from the dataset then combined the ``Monologue'' and ``Others'' classes as one unique class under the name ``Others'' since the ``Monologue'' class does not involve interactions and only the speaker usually exists in the image and the ``Others'' class usually involve everyone around the speaker. In short, our experiments were focused on three objective classes, namely, ``Line-of-Sight Entities'', ``Photographer'', and ``Others''. The statistics of the addressee classes are summarized in Table \ref{tab:table2}.
\begin{table}[t!]
  \centering 
\small
    \begin{tabular}{c|c|c}
      \textbf{} & \textbf{No. of Utterances} & \textbf{Percentage (\%)} \\
      \hline
      Line-of-Sight Entities & 313,079 & 50.86\\
      Photographer & 87,373 & 14.16\\
      Others & 215,058 & 34.94\\
      \hline
      \textbf{Total} & \textbf{615,510} & \\ 
    \end{tabular}
    \caption{Addressee Class Statistics}
    \label{tab:table2}
\end{table}
\smallskip
The classes in our dataset are highly imbalanced. Specifically, the utterances belonging to the ``Line-of-Sight Entities'' class greatly outnumber that of the ``Photographer'' class. To mitigate the problem caused by this imbalance, we applied the concept of cost-sensitive learning introduced by Ganganwar \shortcite{Ganganwar2012} during the training phase. We added class weights calculated based on the absolute number of utterances in each class to reduce the loss of minority classes. This is because if we use the imbalance dataset as it is, a classifier may learn to ignore minor classes and label everything as a majority class. Without cost-sensitive learning, the learned classifier may increase the training set accuracy but decrease overall accuracy. 

The proposed model was implemented using Keras~\footnote{\url{https://github.com/keras-team/keras}} with TensorFlow backend. There were 369,306 utterances and corresponding images used for training; 123,102 for testing and the remaining 123,102 as the validation set for adjusting the classifier. We used GloVe embedding $M_e$ of dimension 100D pre-trained on Wikipedia 2014 and the Gigaword 5 corpus \cite{Pennington14glove:global}. The network was trained using the stochastic gradient descent algorithm. The learning rate was set to 0.001 and the batch size was set to 64. For fair comparison, the same experimental protocol was applied to all experiments.

\subsection{Experimental Results}
\begin{figure}[t!]
\centering
\includegraphics[width=7cm, clip]{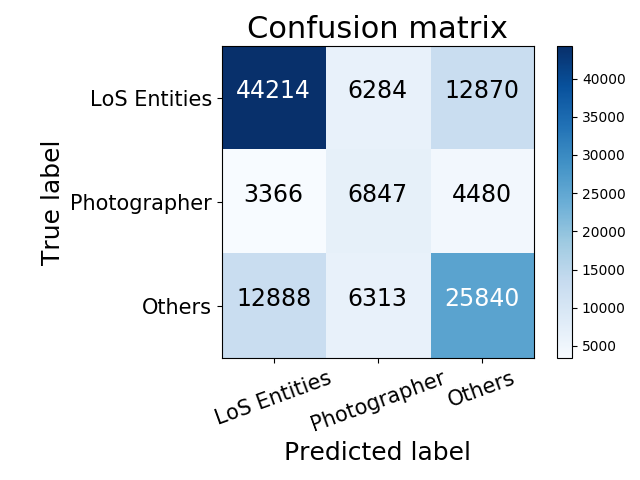}
\caption{Confusion Matrix for Using Our Model on ARVSU Dataset. Each cell shows number of utterances. The term ``LoS Entities" corresponds to "Line-of-Sight Entities" class. Average accuracy was 62.5\%}
\label{fig:confusion_matrix}
\vskip\baselineskip
\end{figure}

We compared our proposed model against the two unimodal recognition models for addressee recognition, as shown in Table \ref{tab:table3}. The evaluation metric used in our experiments was classification accuracy; the higher the accuracy, the better the model’s performance. We report the results at the epoch producing the best validation accuracy. 
\begin{table}[t!]
  \centering 
  \small
    \begin{tabular}{c|c}
      \textbf{Model} & \textbf{Accuracy (\%)} \\
      \hline
      Visual information & 54.0\\
      Text-based utterance & 60.7\\
      \textbf{Visual information + Utterance (proposed)} & \textbf{62.5}\\ 
    \end{tabular}
    \caption{Recognition Performance on ARVSU Dataset}
    \label{tab:table3}
\end{table}
\begin{table*}[ht!]
  \centering 
  \small
	  \begin{tabular}{*{10}c}
		\hline
		Experiment& \multicolumn{3}{c}{Line-of-Sight Entities} & \multicolumn{3}{c}{Photographer} & \multicolumn{3}{c}{Others}\\
		\hline
		{} & $Pre.$ & $Rec.$ & $F_1$ & $Pre.$ & $Rec.$ & $F_1$ & $Pre.$ & $Rec.$ & $F_1$\\
		Visual information  & 62.4 & \textbf{73.7}   & 67.6  & 31.1 &  39.1 & 34.6 & 47.0 & 31.1 & 37.4\\
		Text-based utterance &  \textbf{75.2} & 64.2   & 69.3  & 28.3 &  \textbf{51.3} & 36.5 & \textbf{62.4} & \textbf{58.8} & \textbf{60.6}\\
		\textbf{Visual information + Utterance (proposed)} &  73.1 & 69.8 & \textbf{71.4} & \textbf{35.2} & 46.6 & \textbf{40.1} & 59.8 & 57.4 & 58.6\\
		\hline
     \end{tabular}
    \caption{Per-class Evaluation Results}
    \label{tab:table4}
\end{table*}
The proposed model significantly outperformed both unimodal recognition models. This means our model is able to use all modality components in the dataset to distinguish class labels. The results also suggest that utterances are considerably informative compared to the visual information regarding understanding a conversational context. It is also important to note that we achieved superior performance with mere concatenation over the element-wise multiplication suggested by Recasens et al. \shortcite{recasens2015they}. The result can be explained as the visual and utterance-based features significantly differ from each other. To assess the effectiveness of our model more precisely, we also discuss the evaluation for each class. The confusion matrix for the proposed model is depicted in Figure \ref{fig:confusion_matrix}. 
\smallskip
Our proposed model was especially effective in recognizing ``Line-of-Sight Entities'' and ``Photographer'' classes, which are considered particularly important in human-computer interaction. Our model achieved an F-score of 71.4 and 40.1\% on ``Line-of-Sight Entities'' and ``Photographer'' classes, respectively. However, the text-based-utterance model performed the best for the ``Others'' class. We believe that this is a result of the visual information failing to learn the representative features for the monologue or pondering situations. To illustrate the effectiveness of our model, Table \ref{tab:table4} includes the precision, recall, and F-score for each class.

\subsection{Discussion}
	\label{sec:discussion}
\begin{figure}[t!]
\centering
\includegraphics[width=\columnwidth, clip]{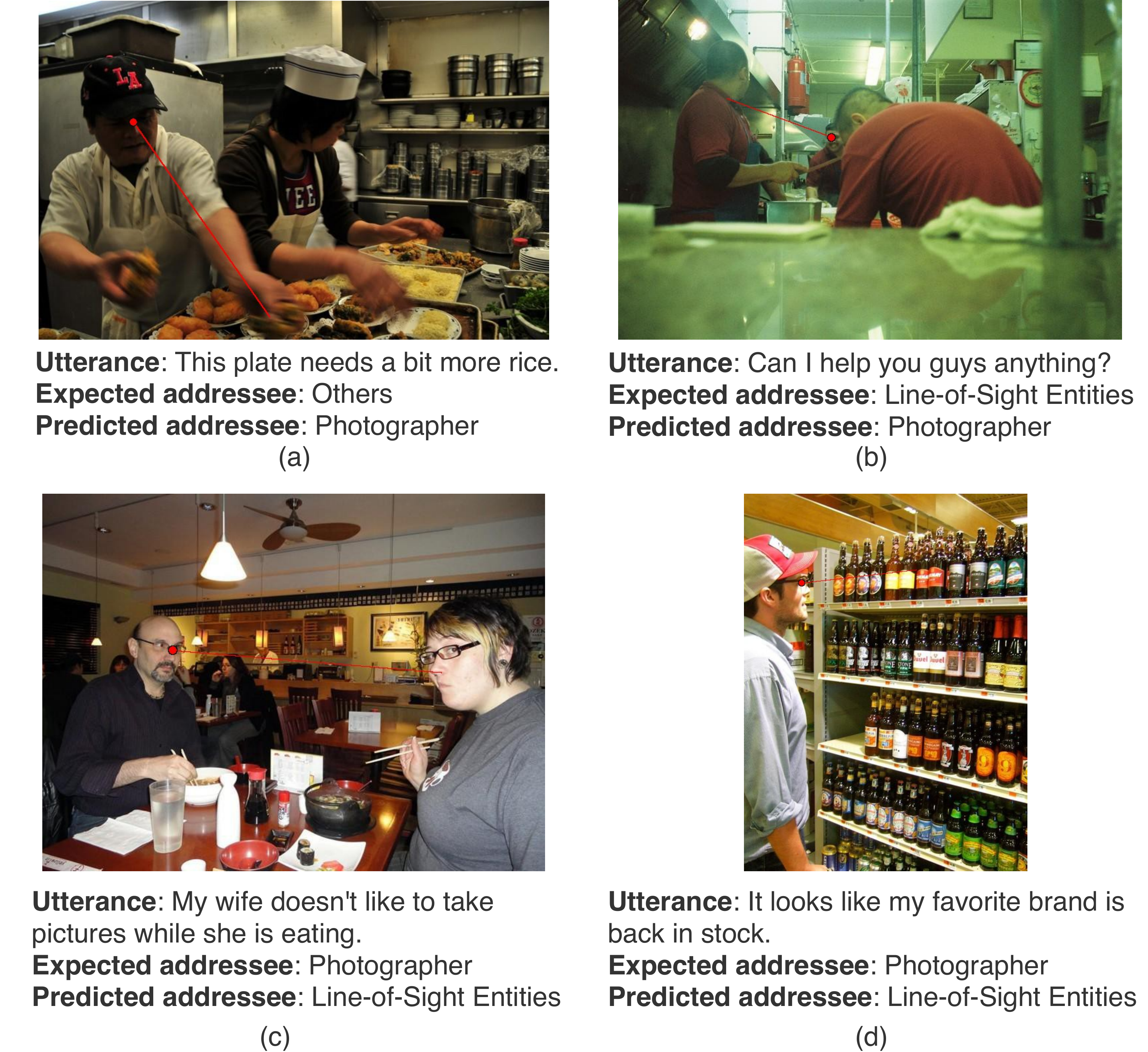}
\caption{Examples Our System Failed to Recognize Properly}
\label{fig:error_case}
\end{figure}

In this section, we present typical examples in which our model failed to recognize the conversational scenario properly. The details of each example are given in Figure \ref{fig:error_case}. In particular, our multi-modal model identified the addressee in case (a) as ``Photographer'' while the expected addressee was ``Others''. In other words, the man seems to talk to the woman next to him based on the inferred scenario from the utterance, but the actual prediction seems to be overwhelmed by the visual features. Another interesting case in which our model could not perform well was when the speaker is looking straight at the photographer but far from the photographer, as presented in case (b). This problem will be included in future work to improve the performance of the model. In cases (c) and (d), we expected our model to recognize the scenario in which a speaker is talking to the photographer, but it could not do so and misclassified the addressee as entities in the line-of-sight. The detailed analysis of the experimental results revealed that our model is unable to distinguish between the ``Photographer'' and ``Others'' classes in many cases. The reason is that the visual information in the scenario of the speaker interacting with a computer agent and monologue seems identical from the computer-agent view. All the examples above are considered difficult situations in real-life conversations, which current intelligent systems are still struggling to solve.

To scale our work for real-world application, an intelligent system must be adapted to streams of visual and verbal information. To obtain an image and associated utterance, we may use the speech recognition system proposed by \cite{7953112}. In their work, speech recognition improved using the features obtained from a randomly selected image frame of video stream during a speech utterance. Because their system selects an image and produces transcribed speech texts, our model can plug into their system and use their outputs as our inputs. We would also need to incorporate human detection as well as head location detection algorithms, which allows the system to automatically estimate speaker positions and head location from original images rather than using annotated positions. In such a system setup, the knowledge of the addressee at time $t-1$ can be used in the model for addressee recognition at time $t$ as well, opening up future research topics. 

\smallskip

\section{Conclusion}
	\label{sec:conclusion}
    Determining who is being addressed in a conversational situation involving a computer agent is crucial for intelligent systems to recognize real-life human social scenes. We introduced the new large mock dataset ARVSU of social visual scenes with annotated utterances. We also proposed a deep learning model using multiple processing modalities to solve the problem of addressee recognition on ARVSU. Our model is able to learn the representative features for different addressee objectives from multi-modal input features, including visual features and text-based utterance features.

Our error analysis revealed that the visual information in the scenario of the speaker interacting with a computer agent and monologue frequently seems identical from the computer-agent view. For future work, to solve this problem, we will take the distance between computer agent and speaker into account. This should allow us to reject monologue cases in which the distance is greater than a specific threshold to prevent misrecognition as a human-computer interaction case. We believe that overcoming these problems will be a stepping stone for future intelligent systems to fully recognize real human social scenes. 
    
\section*{Acknowledgments}
We thank the anonymous reviewers for their careful reading of our manuscript and their many insightful comments and suggestions.

\bibliographystyle{named}
\bibliography{ijcai18}

\end{document}